\documentclass{article}
\usepackage{spconf,amsmath,amssymb,graphicx,hyperref}
\usepackage{multirow,makecell,array,adjustbox}
\usepackage{cleveref}  % after hyperref
\usepackage{xcolor}
\usepackage{pifont}% http://ctan.org/pkg/pifont
\usepackage{orcidlink}
\usepackage{cite}

\newcommand{\cmark}{\ding{51}}%
\newcommand{\xmark}{\ding{55}}%

\definecolor{figblue}{HTML}{1F77B4}
\definecolor{figorange}{HTML}{FF7F0E}
\definecolor{figpurple}{HTML}{9467BD}

\renewcommand{\paragraph}[1]{\par\smallskip\noindent\textbf{#1}\ }

\makeatletter
\long\def\@makecaption#1#2{%
 \vskip 4pt
 \setbox\@tempboxa\hbox{#1. #2}%
 \ifdim \wd\@tempboxa >\hsize #1. #2\par \else \hbox to\hsize{\hfil\box\@tempboxa\hfil}\fi}

\def\thebibliography#1{\section{References}\list
 {[\arabic{enumi}]}{\settowidth\labelwidth{[#1]}\leftmargin\labelwidth
 \advance\leftmargin\labelsep
 \itemsep 0pt \parsep 0pt \topsep 2pt
 \usecounter{enumi}}
 \def\newblock{\hskip .11em plus .33em minus .07em}
 \sloppy\clubpenalty4000\widowpenalty4000
 \sfcode`\.=1000\relax}
\makeatother

\title{One Mean Log-Mel Vector per Phoneme: \\ Simple and Fast Text Injection for Speech Recognition}

\name{
Albert Zeyer$^{1}$ \orcidlink{0000-0002-6655-671X}\thanks{Albert Zeyer was at AppTek during part of this work; now at NVIDIA.},
Davyd Naveriani$^{1}$ \orcidlink{0009-0006-8718-2995},
Dorian Koch$^{1}$ \orcidlink{0009-0001-1273-3514},
Ralf Schlüter$^{1,2}$ \orcidlink{0000-0003-2839-9247},
Hermann Ney$^{1,2}$ \orcidlink{0000-0002-7694-8370}
}

\address{$^{1}$Machine Learning and Human Language Technology Group, RWTH Aachen University, Germany \\
$^{2}$AppTek GmbH, Aachen, Germany\\
\{zeyer,schlueter,ney\}@ml.rwth-aachen.de,\{davyd.naveriani,dorian.koch\}@rwth-aachen.de
}

\begin{document}
\ninept  % 9 pt, the smallest the ICASSP kit allows ("no smaller than nine point type throughout the paper")
\maketitle
\begin{abstract}
Text injection trains speech recognition models on unpaired text alongside paired audio.
We feed the speech encoder synthetic input derived from text
so the whole model learns from it.
Previously, we synthesized this input with a frozen text-to-speech (TTS) model.
Here, a simple lookup table replaces the TTS model:
a mean log-mel vector (or one-hot encoding) per phoneme,
repeated for a random duration drawn from alignment statistics,
with silence between words.
Training is 1.8 times faster than with the neural TTS,
achieving comparable word error rates.
With an external language model, our system reaches
1.36\,/\,2.90\,/\,1.52\,/\,3.24\,\% WER on LibriSpeech \{dev,test\}-\{clean,other\},
to our knowledge state-of-the-art using only LibriSpeech paired data, text, and lexicon.
We analyze components of synthetic input
(durations, acoustic features),
compare different HMM aligners,
evaluate scaling on the 25\,000\,h Loquacious dataset,
and show that full-encoder injection yields strong relative improvements compared to partial-encoder and decoder methods.
The code of the whole pipeline behind every number in this paper is public:
\href{https://github.com/rwth-i6/returnn-experiments/tree/master/2026-text-injection}{github.com/rwth-i6/returnn-experiments/tree/master/2026-text-injection}
\end{abstract}
\begin{keywords}
speech recognition, text injection, unpaired text, pseudo speech, forced alignment
\end{keywords}

\begin{figure}[tb]
  \centering
  \includegraphics[width=\columnwidth]{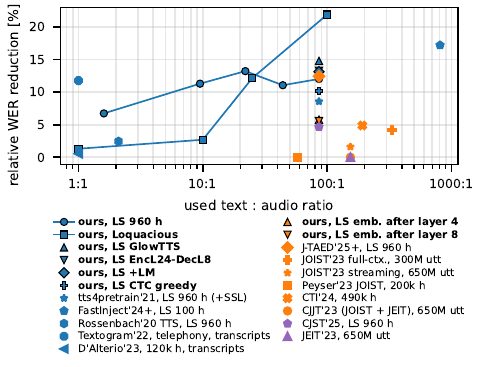}
  \caption{\textbf{Relative WER reduction vs.\ used text : audio ratio}:
  ours (\Cref{tab:ls-text,tab:loq-scale}) and the literature with in-domain text, against the same model without text
  \cite{tang2025jtaed,zhou24cjst,chen2021tts4pretrain,deng2024fastinject,rossenbach:ICASSP2020,thomas2022textogram,dalterio2023tts,sainath2023joist,meng2023jeit,peyser2023ssl,meng2024cti}.
  Color: text enters the whole encoder (\textcolor{figblue}{blue}), a shared upper encoder only (\textcolor{figorange}{orange}), or the decoder only (\textcolor{figpurple}{purple}); $+$: explicit modality-matching loss.
  WER on test-other, Loquacious test, or the general voice-search set.}
  \label{fig:ratio-gain}
\end{figure}

\section{Introduction}

Text without audio exists in far larger amounts than transcribed audio.
Text corpora for language models exceed $10^{13}$ words \cite{dubey2024llama3},
whereas the largest weakly transcribed speech collection (680\,000\,h) \cite{radford2023whisper} contains fewer than $10^{10}$ words.
Even at industrial scale,
transcribed speech is a small fraction of the text:
90\,000\,h (about $10^9$ words) against 28 billion text sentences \cite{zhang2023usm}.
On the scale of this paper,
the 960\,h of LibriSpeech carry 9.4\,M words,
while its text corpus has 800\,M \cite{panayotov2015librispeech}.
Traditionally, text is used for an external language model (LM) in search \cite{prabhavalkar2023end}.
Text injection directly trains the end-to-end model on unpaired text \cite{chen2021tts4pretrain,sainath2023joist,meng2023jeit}.
We inject synthetic input derived from text at the speech encoder input,
training the whole model.
Previously \cite{Zeyer2026TextUtilPseudoSpeechEnc},
a frozen TTS model synthesized this input \cite{bataev2023textonly},
yielding consistent gains but requiring TTS pretraining and dominating training time.
Here, a per-phoneme log-mel table replaces the TTS model.
The input enters where audio features enter,
leaving the model, losses, and search unchanged.

Text injection integrates unpaired text into ASR via decoder-side LMs \cite{wang20222prefix-lm,meng2023jeit,mittal2023insitu,gu2024lm,koch2025dlm}
or encoder-side modality matching,
either explicit via specific alignment losses \cite{chung-etal-2021-splat,chen2022maestro,wang2023understandingsharedrepr,zhou24cjst,gaur24astra}
or implicit via shared dynamics \cite{bapna2021slamunifiedencoderspeech,bapna2022mslammassivelymultilingualjoint,tang-etal-2022-unified,thomas2022textogram,ao-etal-2022-speecht5,zhang2022speechut,sainath2023joist,Sunder2025nonautoregressive}.
While implicit matching generally outperforms explicit matching, TTS synthesis significantly outperforms both \cite{Zeyer2026TextUtilPseudoSpeechEnc,rossenbach:ICASSP2020,dalterio2023tts,gu2024lm,koch2025dlm}.
However, neural TTS introduces severe system complexity and computational expense.
To overcome this, we propose a simplified synthetic acoustic approach, replacing the neural TTS with a simple lookup table, allowing us to easily scale down durations and accelerate ASR training.

The closest approach is the textogram family \cite{thomas2022textogram,dalterio2023tts,huang2023ustr,deng2024fastinject,tang2025jtaed,Sunder2025nonautoregressive},
where text enters the encoder as a symbol sequence, often masked.
Our input is instead acoustic:
a mean log-mel vector per phoneme,
held for a duration drawn from alignment statistics.
It takes the audio path with SpecAugment,
requiring no text encoder, masking, or extra losses.
Alignment-derived durations perform best (\Cref{sec:results}), though simpler character textograms (fixed 4 frames) still retain 60\,\% of our gain (\Cref{tab:ls-representation}).
Previous textogram results report large relative gains but rely on weak baselines (5.5--28\,\% WER) or out-of-domain text \cite{tang2025jtaed,deng2024fastinject,huang2023ustr,Sunder2025nonautoregressive,thomas2022textogram,dalterio2023tts}.
Only \cite{tang2025jtaed} shares our strict condition of using solely matched-domain text from scratch.

Our contributions are fourfold:
(1) replacing neural TTS with a synthesizer-free lookup table, accelerating training by 1.8$\times$;
(2) achieving a 12\,\% relative WER reduction on LibriSpeech, reaching state-of-the-art for the constrained data case;
(3) demonstrating through ablations and textogram comparisons that simple duration and acoustic models suffice for strong performance; and
(4) comparing across the literature, we show that full-encoder injection yields highly competitive relative improvements over partial-encoder or decoder methods.

% Method overview, one column (figures/method/README.md; without interpolation: method-overview-S7-nolerp.pdf).
% DISABLED 2026-09-24: with it the paper exceeds the page limit (about 95 mm of references spill onto page 6).
% Re-enable by uncommenting, together with the \Cref in Sec. 2.
% \begin{figure}[t]
%   \centering
%   \includegraphics[width=\columnwidth]{figures/method-overview-S7.pdf}
%   \caption{\textbf{Text injection with a per-phoneme log-mel table.}
%   Offline, a forced alignment of the paired audio gives each phoneme a mean log-mel vector $\mu$ and a median duration $m$.
%   A text sentence is mapped to phonemes, each held for a sampled duration $d$ with its vector $\mu$ interpolated towards the next.
%   The result enters the unchanged model like real audio.}
%   \label{fig:method}
% \end{figure}

\section{Method}
\label{sec:method}

% with the method figure (fig:method) re-enabled: "...of the paired audio (\Cref{fig:method}),"
From a phoneme alignment of the paired audio,
we extract a mean log-mel vector and median duration $m$ per phoneme.
To inject text, we map sentences to phonemes via a lexicon.
Each phoneme duration is sampled as $\operatorname{round}(m \cdot s \cdot \exp(\sigma \varepsilon))$,
$\varepsilon \sim \mathcal{N}(0, 1)$,
with scale $s$ and jitter $\sigma$ (\Cref{tab:ls-durations}).
We insert silence between words,
repeat vectors for their duration with linear interpolation at boundaries,
and feed this to the front-end exactly like real audio,
applying SpecAugment.
Audio and text sentences are mixed randomly in batches,
using the same losses.

\section{Experiments}
\label{sec:experiments}

% the tables are auto-generated (scripts/sync_tables.sh), see README.

\subsection{Setup}
\label{sec:setup}

We use LibriSpeech \cite{panayotov2015librispeech}
(960\,h paired, 800M-word LM corpus for injection)
with a pronunciation lexicon,
and Loquacious \cite{parcollet2025loquacious}
(25\,000\,h English from five sources).
Loquacious subsets include small (250\,h), medium (2\,500\,h), large (25\,000\,h),
and a custom 1\,000\,h subset.
We use the large subset transcripts for text injection experiments.

Models use a 16-layer Conformer encoder and 6-layer Transformer decoder
(dimension 1024, 560M parameters; EncL24-DecL8 uses 24 and 8 layers).
We train with CTC+AED on 10k sentence pieces
from 80-dim log-mel features with a 6x subsampling front-end,
using the Muon optimizer \cite{jordan2024muon},
packed tensors and CUDA graphs on 4 GH200 GPUs.
External LMs are standard auto-regressive Transformers.
The larger LibriSpeech model also uses diffusion \cite{naveriani2026:difflm}
and denoising \cite{koch2025dlm} LMs.

For LibriSpeech we take the public phoneme alignments \cite{lugosch2019slu,lugosch2019librispeechalignments}
of the Montreal Forced Aligner (MFA) \cite{mcauliffe2017mfa}, a speaker-adapted triphone Gaussian-mixture HMM (GM-HMM) on MFCC features.
Its pretrained HMM was itself trained strictly on LibriSpeech, so no external or foreign audio enters the tables;
Loquacious is aligned with the same model.
MFA relies on a complex pipeline,
so we also evaluate a simple, self-contained aligner (\Cref{tab:ls-table-source}):
a monophone HMM with three states per phoneme and one diagonal Gaussian per state,
trained on the ASR log-mel features in RETURNN \cite{zeyer2018:returnn}.
Its state means directly form the log-mel table.

Decoding is single-pass without rescoring \cite{prabhavalkar2023end,koch2025dlm,zhou2021:segmental-transducer}.
CTC, CTC+AED, and CTC+LM use time-synchronous search,
while CTC+AED+LM uses label-sync.\ search over CTC prefix scores.

\subsection{Results}
\label{sec:results}

\paragraph{Injection vs.\ baselines.}
\Cref{tab:ls-main} compares text injection to audio-only baselines on LibriSpeech.
The table reduces WER by 12\,\% relative on test-other (4.16 to 3.66).
The gain persists with an external LM (3.38 to 2.93)
and is not recovered by simply doubling audio epochs.
The larger model sees similar gains (3.71 to 3.26, and 2.90 with LM).
The denoising LM slightly outperforms the std.~LM on test-other (3.22 vs.\ 3.24).
The diffusion LM underperforms CTC+LM (3.81 vs.\ 3.33)
but is much faster,
requiring only a single non-autoregressive denoising step \cite{naveriani2026:difflm}.
Compared to the GlowTTS injection \cite{Zeyer2026TextUtilPseudoSpeechEnc},
our lookup table matches the WER while training 1.8 times faster
(130\,h vs.\ 235\,h on 4 GPUs in an earlier implementation).
% =====================================================================
% AUTO-GENERATED by scripts/render_tables.py -- DO NOT EDIT THIS FILE.
% Regenerate from (edit these, then re-run render_tables.py / sync_tables_preview.sh):
%   tables-data/ls-main.data.json    -- the numbers (from the Sisyphus graph)
%   tables-spec/ls-main.spec.json     -- columns, units, layout
%   tables-spec/ls-main.caption.tex   -- caption text
% =====================================================================
% data source (Sisyphus recipe): utils.table_data.WriteTableDataJob (flat)
% per-row comments below map columns -> the registered output name the number comes from
\begin{table}[tb]
  \centering
  \caption{\textbf{Text-injection vs.\ baselines},
LibriSpeech 960h + LM text,
with and without an external LM
(std.: standard auto-regressive,
diff.: diffusion \cite{naveriani2026:difflm},
den.: denoising \cite{koch2025dlm}),
and training time on 4 GH200 GPUs.
Mean log-mel: tables from MFA phoneme alignment.
None (240k updates): twice the epochs.
TTS: GlowTTS text-injection, WERs from \cite{Zeyer2026TextUtilPseudoSpeechEnc} (earlier implementation, single GPU).}

  \label{tab:ls-main}
  \begin{adjustbox}{max width=\linewidth}
  \begin{tabular}{|c|l|c|c||c|c|c|c|c|c|}
    \hline
    \multirow{3}{*}{\textbf{\rotatebox{90}{Model}}} & \multirow{3}{*}{\textbf{\makecell[l]{Text \\ injection}}} & \multirow{3}{*}{\textbf{\makecell{Up- \\ dates \\ {\normalfont [k]}}}} & \multirow{3}{*}{\textbf{\makecell{Train \\ time \\ {\normalfont [h]}}}} & \multicolumn{2}{c|}{\textbf{CTC+}} & \multicolumn{4}{c|}{\textbf{WER {\normalfont [\%]}}} \\
     &  &  &  & \multirow{2}{*}{\textbf{AED}} & \multirow{2}{*}{\textbf{LM}} & \multicolumn{2}{c|}{\textbf{dev}} & \multicolumn{2}{c|}{\textbf{test}} \\
     &  &  &  &  &  & \textbf{clean} & \textbf{other} & \textbf{clean} & \textbf{other} \\ \hline\hline
    \multirow{10}{*}{\makecell[l]{\rotatebox{90}{\footnotesize\makecell{EncL16 \\ DecL6}}}} & \multirow{5}{*}{none} & \multirow{4}{*}{$120$} & \multirow{4}{*}{$20$} & \xmark & \multirow{2}{*}{\xmark} & $2.11$ & $4.42$ & $2.27$ & $4.83$ \\
    \cline{5-5}\cline{7-10}
     &  &  &  & \cmark &  & $1.78$ & $3.85$ & $1.96$ & $4.16$ \\
    \cline{5-10}
     &  &  &  & \xmark & \multirow{2}{*}{std.} & $1.78$ & $3.68$ & $1.93$ & $4.01$ \\
    \cline{5-5}\cline{7-10}
     &  &  &  & \multirow{3}{*}{\cmark} &  & $1.69$ & $3.38$ & $1.84$ & $3.73$ \\
    \cline{3-4}\cline{6-10}
     &  & $240$ & $40$ &  & \multirow{4}{*}{\xmark} & $1.81$ & $3.71$ & $2.01$ & $4.14$ \\
    \cline{2-4}\cline{7-10}
     & TTS \cite{Zeyer2026TextUtilPseudoSpeechEnc} & \multicolumn{2}{c||}{single GPU} &  &  & 1.57 & 3.53 & 1.73 & 3.64 \\
    \cline{2-5}\cline{7-10}
     & \multirow{4}{*}{\makecell[l]{mean \\ log-mel}} & \multirow{4}{*}{$246$} & \multirow{4}{*}{$44$} & \xmark &  & $1.78$ & $3.95$ & $1.92$ & $4.34$ \\
    \cline{5-5}\cline{7-10}
     &  &  &  & \cmark &  & $1.49$ & $3.40$ & $1.65$ & $3.66$ \\
    \cline{5-10}
     &  &  &  & \xmark & \multirow{2}{*}{std.} & $1.45$ & $3.05$ & $1.56$ & $3.44$ \\
    \cline{5-5}\cline{7-10}
     &  &  &  & \multirow{2}{*}{\cmark} &  & $1.36$ & $2.93$ & $1.56$ & $3.24$ \\
    \cline{1-4}\cline{6-10}
    \multirow{7}{*}{\makecell[l]{\rotatebox{90}{\footnotesize\makecell{EncL24 \\ DecL8}}}} & none & $120$ & $28$ &  & \multirow{3}{*}{\xmark} & $1.78$ & $3.71$ & $1.92$ & $4.07$ \\
    \cline{2-5}\cline{7-10}
     & \multirow{6}{*}{\makecell[l]{mean \\ log-mel}} & \multirow{6}{*}{$246$} & \multirow{6}{*}{$65$} & \xmark &  & $1.75$ & $3.66$ & $1.84$ & $3.99$ \\
    \cline{5-5}\cline{7-10}
     &  &  &  & \cmark &  & $1.51$ & $3.26$ & $1.60$ & $3.53$ \\
    \cline{5-10}
     &  &  &  & \xmark & \multirow{2}{*}{std.} & $1.45$ & $3.02$ & $1.61$ & $3.33$ \\
    \cline{5-5}\cline{7-10}
     &  &  &  & \cmark &  & $1.36$ & $2.90$ & $1.52$ & $3.24$ \\
    \cline{5-10}
     &  &  &  & \xmark & diff. & $1.63$ & $3.48$ & $1.77$ & $3.81$ \\
    \cline{5-10}
     &  &  &  & \cmark & den. & $1.40$ & $2.89$ & $1.62$ & $3.22$ \\
    \hline
  \end{tabular}
  \end{adjustbox}
\end{table}

\paragraph{Aligner.}
\Cref{tab:ls-table-source} evaluates our self-contained aligner (\Cref{sec:setup}).
It retains most of the gain (3.46 vs.\ 3.40 WER),
proving a single Gaussian per state is sufficient.
% =====================================================================
% AUTO-GENERATED by scripts/render_tables.py -- DO NOT EDIT THIS FILE.
% Regenerate from (edit these, then re-run render_tables.py / sync_tables_preview.sh):
%   tables-data/ls-table-source.data.json    -- the numbers (from the Sisyphus graph)
%   tables-spec/ls-table-source.spec.json     -- columns, units, layout
%   tables-spec/ls-table-source.caption.tex   -- caption text
% =====================================================================
% data source (Sisyphus recipe): utils.table_data.WriteTableDataJob (flat)
% per-row comments below map columns -> the registered output name the number comes from
\begin{table}[tb]
  \centering
  \caption{\textbf{Aligner:} replacing the MFA phoneme alignment
with our own single-Gaussian HMM on ASR features,
extracting one mean log-mel vector per phoneme or HMM state (3 per phoneme),
and duration statistics,
and evaluating text injection on LibriSpeech.
}

  \label{tab:ls-table-source}
  \begin{adjustbox}{max width=\linewidth}
  \begin{tabular}{|l|c|l|l|c|c|c|c|}
    \hline
    \multirow{3}{*}{\textbf{Aligner}} & \multirow{3}{*}{\textbf{\makecell{Pron. \\ variants}}} & \multirow{3}{*}{\textbf{\makecell[l]{Log-mel \\ table}}} & \multirow{3}{*}{\textbf{\makecell[l]{Vector \\ per}}} & \multicolumn{4}{c|}{\textbf{WER {\normalfont [\%]}}} \\
     &  &  &  & \multicolumn{2}{c|}{\textbf{dev}} & \multicolumn{2}{c|}{\textbf{test}} \\
     &  &  &  & \textbf{clean} & \textbf{other} & \textbf{clean} & \textbf{other} \\ \hline\hline
    \makecell[l]{triphone \\ GM-HMM \\ (via MFA)} & all & \makecell[l]{frame \\ mean} & \multirow{4}{*}{phoneme} & $1.49$ & $3.40$ & $1.65$ & $3.66$ \\
    \cline{1-3}\cline{5-8}
    \multirow{4}{*}{\makecell[l]{monophone \\ single-Gauss. \\ HMM (ours)}} & sampled & \multirow{2}{*}{\makecell[l]{Gauss. \\ means}} &  & $1.58$ & $3.45$ & $1.73$ & $3.74$ \\
    \cline{2-2}\cline{5-8}
     & \multirow{3}{*}{all} &  &  & $1.51$ & $3.46$ & $1.70$ & $3.70$ \\
    \cline{3-3}\cline{5-8}
     &  & \makecell[l]{frame \\ mean} &  & $1.53$ & $3.43$ & $1.71$ & $3.69$ \\
    \cline{3-8}
     &  & \makecell[l]{Gauss. \\ means} & \makecell[l]{HMM \\ state} & $1.51$ & $3.44$ & $1.64$ & $3.67$ \\
    \hline
  \end{tabular}
  \end{adjustbox}
\end{table}

Directly using subwords failed (no gain over audio-only; not shown in table),
whether resampling a mean vector per subword or sampling real acoustic slices.
Listening to the synthetic input, we found subword acoustics are highly context-dependent,
making simple mean arrays less effective,
whereas phonemes are invariant enough for a single mean vector.
This requires further investigation.

\paragraph{Durations.}
\Cref{tab:ls-durations} varies the duration model.
The log-mel table is robust to temporal compression.
Scaling durations down to 0.7x (5.9 frames per phoneme vs.\ 8.4 in the alignment) performs best (3.66 on test-other).
Even collapsing durations to a constant 6 frames per phoneme (60\,ms) incurs only a minor penalty (3.73).
In contrast, neural TTS appears less robust to larger modifications of its generated durations.
While an exact matched comparison is unavailable, randomly scaling GlowTTS \cite{Zeyer2026TextUtilPseudoSpeechEnc} durations by factors sampled uniformly between 0.2 and 0.5 degraded test-other by roughly +0.5\,\% absolute WER.
% =====================================================================
% AUTO-GENERATED by scripts/render_tables.py -- DO NOT EDIT THIS FILE.
% Regenerate from (edit these, then re-run render_tables.py / sync_tables_preview.sh):
%   tables-data/ls-durations.data.json    -- the numbers (from the Sisyphus graph)
%   tables-spec/ls-durations.spec.json     -- columns, units, layout
%   tables-spec/ls-durations.caption.tex   -- caption text
% =====================================================================
% data source (Sisyphus recipe): utils.table_data.WriteTableDataJob (flat)
% per-row comments below map columns -> the registered output name the number comes from
\begin{table}[tb]
  \centering
  \caption{\textbf{Durations}
$d = \mathrm{round}(m \cdot s \cdot \exp(\sigma \varepsilon))$,
$\varepsilon \sim \mathcal{N}(0,1)$
on LibriSpeech, varying scale $s$ and jitter $\sigma$,
with per-phoneme median $m$
vs one median for all phonemes, uniform range, and fixed count.
Mean dur.: the expected frames (10\,ms) per phoneme under the sampling; the alignment's mean is 8.4.
}

  \label{tab:ls-durations}
  \begin{adjustbox}{max width=\linewidth}
  \begin{tabular}{|l|c|c|c|c|c|c|}
    \hline
    \multirow{3}{*}{\textbf{\makecell[l]{Duration \\ distribution}}} & \multirow{3}{*}{\textbf{\makecell{Scale \\ $s$}}} & \multirow{3}{*}{\textbf{\makecell{Jitter \\ $\sigma$}}} & \multirow{3}{*}{\textbf{\makecell{Mean \\ dur. \\ {\normalfont [frames]}}}} & \multicolumn{2}{c|}{\textbf{WER {\normalfont [\%]}}} & \multirow{3}{*}{\textbf{\makecell{Train \\ time \\ {\normalfont [h]}}}} \\
     &  &  &  & \multicolumn{1}{c|}{\textbf{dev}} & \multicolumn{1}{c|}{\textbf{test}} &  \\
     &  &  &  & \textbf{other} & \textbf{other} &  \\ \hline\hline
    \multirow{6}{*}{\makecell[l]{log-normal, \\ per-phoneme \\ median}} & 0.5 & 0.45 & $4.3$ & $3.52$ & $3.73$ & $44$ \\
    \cline{2-7}
     & \multirow{4}{*}{0.7} & 0.00 & $5.4$ & $3.53$ & $3.79$ & $45$ \\
    \cline{3-7}
     &  & 0.20 & $5.5$ & $3.50$ & $3.72$ & $46$ \\
    \cline{3-7}
     &  & 0.45 & $6.0$ & $3.40$ & $3.66$ & $44$ \\
    \cline{3-7}
     &  & 0.70 & $6.9$ & $3.45$ & $3.74$ & $45$ \\
    \cline{2-7}
     & 1.0 & \multirow{2}{*}{0.45} & $8.5$ & $3.49$ & $3.64$ & $54$ \\
    \cline{1-2}\cline{4-7}
    \makecell[l]{log-normal, \\ one median \\ for all phonemes} & 0.7 &  & $5.4$ & $3.50$ & $3.71$ & $45$ \\
    \hline
    uniform 5 to 10 & \multirow{2}{*}{-} & \multirow{2}{*}{-} & $7.5$ & $3.53$ & $3.78$ & $47$ \\
    \cline{1-1}\cline{4-7}
    fixed 6 &  &  & $6.0$ & $3.47$ & $3.73$ & $44$ \\
    \hline
  \end{tabular}
  \end{adjustbox}
\end{table}

\paragraph{Acoustic features.}
\Cref{tab:ls-representation} compares acoustic features.
The acoustic features matter little:
representing phonemes as trainable embeddings (3.47) or discrete one-hot channels concatenated to zeroed log-mels (3.37) performs similarly to using mean log-mel features (3.40).
Characters with a fixed 4 frames per symbol, the original textogram configuration \cite{thomas2022textogram}, give 3.58: worse than any phoneme row, but still 0.27 better than the baseline (3.85, \Cref{tab:ls-main}).
However, durations matter more:
uniform durations cost the log-mel table 0.13 WER.
Other rendering details have negligible impact (not shown):
removing linear interpolation for hard repeats (3.39 dev-other), removing inter-word silences (3.44), or adding silence at utterance boundaries (3.39) barely deviate from the 3.40 baseline.
% =====================================================================
% AUTO-GENERATED by scripts/render_tables.py -- DO NOT EDIT THIS FILE.
% Regenerate from (edit these, then re-run render_tables.py / sync_tables_preview.sh):
%   tables-data/ls-representation.data.json    -- the numbers (from the Sisyphus graph)
%   tables-spec/ls-representation.spec.json     -- columns, units, layout
%   tables-spec/ls-representation.caption.tex   -- caption text
% =====================================================================
% data source (Sisyphus recipe): utils.table_data.WriteTableDataJob (flat)
% per-row comments below map columns -> the registered output name the number comes from
\begin{table}[tb]
  \centering
  \caption{\textbf{Acoustic features and units:}
varying the acoustic features
(mean log-mel, trainable embeddings,
or discrete one-hot channels concatenated to the log-mel),
injected units (phonemes vs.\ characters),
and duration distributions on LibriSpeech.
Seq.\ len.: mean frames (10\,ms) per injected LM sentence; 729 at the alignment's mean durations.
Last row: configuration of original \emph{textogram} \cite{thomas2022textogram}.
%graphemes with fixed 4 frames per symbol.
}

  \label{tab:ls-representation}
  \begin{adjustbox}{max width=\linewidth}
  \begin{tabular}{|l|l|l|c|c|c|}
    \hline
    \multirow{3}{*}{\textbf{\makecell[l]{Acoustic \\ features}}} & \multirow{3}{*}{\textbf{Units}} & \multirow{3}{*}{\textbf{Durations}} & \multirow{3}{*}{\textbf{\makecell{Seq. \\ len. \\ {\normalfont [frames]}}}} & \multicolumn{2}{c|}{\textbf{WER {\normalfont [\%]}}} \\
     &  &  &  & \multicolumn{1}{c|}{\textbf{dev}} & \multicolumn{1}{c|}{\textbf{test}} \\
     &  &  &  & \textbf{other} & \textbf{other} \\ \hline\hline
    \multirow{2}{*}{\makecell[l]{mean \\ log-mel}} & \multirow{6}{*}{phonemes} & \makecell[l]{log-normal \\ per phoneme} & $496$ & $3.40$ & $3.66$ \\
    \cline{3-6}
     &  & \makecell[l]{uniform \\ 5 to 10} & $574$ & $3.53$ & $3.78$ \\
    \cline{1-1}\cline{3-6}
    \multirow{2}{*}{\makecell[l]{trainable \\ log-mel}} &  & \makecell[l]{log-normal \\ per phoneme} & $496$ & $3.47$ & $3.69$ \\
    \cline{3-6}
     &  & \multirow{2}{*}{\makecell[l]{uniform \\ 5 to 10}} & $574$ & $3.48$ & $3.72$ \\
    \cline{1-1}\cline{4-6}
    \multirow{3}{*}{\makecell[l]{extra one-hot \\ channels}} &  &  & $574$ & $3.37$ & $3.69$ \\
    \cline{3-6}
     &  & fixed 6 & $459$ & $3.43$ & $3.64$ \\
    \cline{2-6}
     & characters & fixed 4 & $428$ & $3.58$ & $3.81$ \\
    \hline
  \end{tabular}
  \end{adjustbox}
\end{table}

\paragraph{Data amounts.}
\Cref{tab:ls-text} and \Cref{tab:ls-audio-amount} vary data amounts.
The optimal sampling ratio is one text word per audio transcript word (\textit{Seen} 1:1).
Text diversity is secondary:
10\,\% of the corpus repeated 19 times is only 0.15 WER behind the full corpus.
Crucially, text cannot replace audio:
halving the audio severely degrades WER,
confirming that paired data still drives the performance.
% =====================================================================
% AUTO-GENERATED by scripts/render_tables.py -- DO NOT EDIT THIS FILE.
% Regenerate from (edit these, then re-run render_tables.py / sync_tables_preview.sh):
%   tables-data/ls-text.data.json    -- the numbers (from the Sisyphus graph)
%   tables-spec/ls-text.spec.json     -- columns, units, layout
%   tables-spec/ls-text.caption.tex   -- caption text
% =====================================================================
% data source (Sisyphus recipe): utils.table_data.WriteTableDataJob (flat)
% per-row comments below map columns -> the registered output name the number comes from
\begin{table}[tb]
  \centering
  \caption{\textbf{Amount of text} on LibriSpeech,
varying text corpus size and injection sampling ratio.
\textit{Used text : audio} is the absolute size ratio of the available unpaired text dataset to the paired audio transcripts.
\textit{Seen text : audio} is the relative proportion of text to audio words actually sampled in a training batch on average.
Top row is audio only.
}

  \label{tab:ls-text}
  \begin{adjustbox}{max width=\linewidth}
  \begin{tabular}{|c|c|c|r|r|c|c|}
    \hline
    \textbf{\makecell{Seen}} & \multirow{3}{*}{\textbf{\makecell{LM \\ text \\ used}}} & \textbf{\makecell{Used}} & \multirow{3}{*}{\textbf{\makecell{Passes \\ over \\ text}}} & \multirow{3}{*}{\textbf{\makecell{Up- \\ dates \\ {\normalfont [k]}}}} & \multicolumn{2}{c|}{\textbf{WER {\normalfont [\%]}}} \\
    \textbf{\makecell{text :}} &  & \textbf{\makecell{text :}} &  &  & \multicolumn{1}{c|}{\textbf{dev}} & \multicolumn{1}{c|}{\textbf{test}} \\
    \textbf{\makecell{audio \\ {\normalfont (words)}}} &  & \textbf{\makecell{audio \\ {\normalfont (words)}}} &  &  & \textbf{other} & \textbf{other} \\ \hline\hline
    0:1 & - & - & 0 & $240$ & $3.71$ & $4.14$ \\
    \hline
    1:4 & \multirow{3}{*}{100\%} & \multirow{3}{*}{86:1} & 0.5 & $170$ & $3.56$ & $3.99$ \\
    \cline{1-1}\cline{4-7}
    1:2 &  &  & 1 & $171$ & $3.54$ & $3.81$ \\
    \cline{1-1}\cline{4-7}
    \multirow{4}{*}{1:1} &  &  & 2 & $246$ & $3.40$ & $3.66$ \\
    \cline{2-7}
     & 50\% & 44:1 & 4 & $246$ & $3.55$ & $3.70$ \\
    \cline{2-7}
     & 25\% & 22:1 & 8 & $252$ & $3.54$ & $3.61$ \\
    \cline{2-7}
     & 10\% & 9.5:1 & 19 & $256$ & $3.55$ & $3.69$ \\
    \hline
    1.6:1 & 0.65\% & 1.6:1 & 152 & $347$ & $3.65$ & $3.88$ \\
    \hline
    2:1 & 100\% & 86:1 & 4 & $498$ & $3.57$ & $3.65$ \\
    \hline
  \end{tabular}
  \end{adjustbox}
\end{table}

% =====================================================================
% AUTO-GENERATED by scripts/render_tables.py -- DO NOT EDIT THIS FILE.
% Regenerate from (edit these, then re-run render_tables.py / sync_tables_preview.sh):
%   tables-data/ls-audio-amount.data.json    -- the numbers (from the Sisyphus graph)
%   tables-spec/ls-audio-amount.spec.json     -- columns, units, layout
%   tables-spec/ls-audio-amount.caption.tex   -- caption text
% =====================================================================
% data source (Sisyphus recipe): utils.table_data.WriteTableDataJob (flat)
% per-row comments below map columns -> the registered output name the number comes from
\begin{table}[tb]
  \centering
  \caption{\textbf{Amount of paired audio}
of LibriSpeech
with the full LM text corpus.}

  \label{tab:ls-audio-amount}
  \begin{adjustbox}{max width=\linewidth}
  \begin{tabular}{|r|r|c|c|r|r|c|c|}
    \hline
    \multirow{3}{*}{\textbf{\makecell{Audio \\ {\normalfont [h]}}}} & \multirow{3}{*}{\textbf{\makecell{Audio \\ pass.}}} & \multirow{3}{*}{\textbf{\makecell{Seen \\ text : \\ audio}}} & \multirow{3}{*}{\textbf{\makecell{Used \\ text : \\ audio}}} & \multirow{3}{*}{\textbf{\makecell{Text \\ pass.}}} & \multirow{3}{*}{\textbf{\makecell{Up- \\ dates \\ {\normalfont [k]}}}} & \multicolumn{2}{c|}{\textbf{WER {\normalfont [\%]}}} \\
     &  &  &  &  &  & \multicolumn{1}{c|}{\textbf{dev}} & \multicolumn{1}{c|}{\textbf{test}} \\
     &  &  &  &  &  & \textbf{other} & \textbf{other} \\ \hline\hline
    \multirow{2}{*}{960} & $152$ & 1:1 & \multirow{2}{*}{86:1} & $2.0$ & $246$ & $3.40$ & $3.66$ \\
    \cline{2-3}\cline{5-8}
     & $38$ & 7:1 &  & $3.5$ & $429$ & $4.75$ & $4.91$ \\
    \hline
    480 & \multirow{5}{*}{$152$} & 3:1 & 170:1 & $3.0$ & $369$ & $4.20$ & $4.33$ \\
    \cline{1-1}\cline{3-8}
    \multirow{2}{*}{240} &  & 7:1 & \multirow{2}{*}{350:1} & $3.5$ & $429$ & $5.06$ & $5.17$ \\
    \cline{3-3}\cline{5-8}
     &  & 4:1 &  & $1.9$ & $233$ & $5.16$ & $5.05$ \\
    \cline{1-1}\cline{3-8}
    \multirow{2}{*}{96} &  & 19:1 & \multirow{2}{*}{860:1} & $3.9$ & $473$ & $6.87$ & $7.11$ \\
    \cline{3-3}\cline{5-8}
     &  & 11:1 &  & $2.0$ & $239$ & $6.66$ & $6.83$ \\
    \hline
    0 & $0$ & 1:0 & 1:0 & $4.0$ & $485$ & \multicolumn{2}{c|}{$>$100} \\
    \hline
  \end{tabular}
  \end{adjustbox}
\end{table}

\paragraph{Batch mixing.}
\Cref{tab:ls-mixing} shows that mixing text and audio in the same batch
outperforms separate batches per modality by 1.4\,\% absolute WER for the TTS injection;
for the log-mel table, random order beats a fixed interleave by 0.08.
% =====================================================================
% AUTO-GENERATED by scripts/render_tables.py -- DO NOT EDIT THIS FILE.
% Regenerate from (edit these, then re-run render_tables.py / sync_tables_preview.sh):
%   tables-data/ls-mixing.data.json    -- the numbers (from the Sisyphus graph)
%   tables-spec/ls-mixing.spec.json     -- columns, units, layout
%   tables-spec/ls-mixing.caption.tex   -- caption text
% =====================================================================
% data source (Sisyphus recipe): utils.table_data.WriteTableDataJob (flat)
% per-row comments below map columns -> the registered output name the number comes from
\begin{table}[tb]
  \centering
  \caption{\textbf{Batch mixing}
on LibriSpeech:
separate batches per modality (alternating),
vs.\ randomly mixed in one batch.
}

  \label{tab:ls-mixing}
  \begin{adjustbox}{max width=\linewidth}
  \begin{tabular}{|l|l|r|c|c|}
    \hline
    \multirow{3}{*}{\textbf{Injection}} & \multirow{3}{*}{\textbf{Batches}} & \multirow{3}{*}{\textbf{\makecell{Up- \\ dates \\ {\normalfont [k]}}}} & \multicolumn{2}{c|}{\textbf{WER {\normalfont [\%]}}} \\
     &  &  & \multicolumn{1}{c|}{\textbf{dev}} & \multicolumn{1}{c|}{\textbf{test}} \\
     &  &  & \textbf{other} & \textbf{other} \\ \hline\hline
    \multirow{3}{*}{TTS} & \makecell[l]{separate batches \\ (audio or text)} & $977$ & $4.92$ & $5.18$ \\
    \cline{2-5}
     & \makecell[l]{mixed batches, \\ fixed ratio} & $805$ & $3.55$ & $3.69$ \\
    \cline{2-5}
     & \makecell[l]{mixed batches, \\ random order} & $805$ & $3.56$ & $3.76$ \\
    \hline
    \multirow{2}{*}{mean log-mel} & \makecell[l]{mixed batches, \\ fixed ratio} & $246$ & $3.48$ & $3.77$ \\
    \cline{2-5}
     & \makecell[l]{mixed batches, \\ random order} & $246$ & $3.40$ & $3.66$ \\
    \hline
  \end{tabular}
  \end{adjustbox}
\end{table}

\paragraph{Loquacious.}
\Cref{tab:loq-scale} repeats the comparison on Loquacious at different scales.
The relative gain depends strongly on the text-to-audio ratio and the absolute amount of paired data.
LibriSpeech achieved a 12\,\% relative gain using an 86:1 text-to-audio ratio (960\,h audio vs.\ an 800M-word text corpus).
On Loquacious, we scale the paired audio while keeping the injected text fixed (the 25\,000\,h transcripts).
The relative gain shrinks predictably as the text-to-audio ratio drops:
24\,\% at 250\,h (100:1 ratio), 14\,\% at 1\,000\,h (25:1), 8\,\% at 2\,500\,h (10:1), and 3\,\% at 25\,000\,h (1:1 ratio).
These gains are against the audio-only baseline at the same number of audio passes.
The controls with twice the passes show that the medium and large baselines are not converged:
there, twice the audio passes give as much as the injection at a similar update count,
while on medium1k they give nothing (8.07 vs.\ 8.04 on dev) and on LibriSpeech twice the epochs recover only 0.14 of the 0.45 gain (\Cref{tab:ls-main}).
Crucially, at matched compute, text injection matches or exceeds extended audio training, providing an efficient alternative without extra audio passes.
% =====================================================================
% AUTO-GENERATED by scripts/render_tables.py -- DO NOT EDIT THIS FILE.
% Regenerate from (edit these, then re-run render_tables.py / sync_tables_preview.sh):
%   tables-data/loq-scale.data.json    -- the numbers (from the Sisyphus graph)
%   tables-spec/loq-scale.spec.json     -- columns, units, layout
%   tables-spec/loq-scale.caption.tex   -- caption text
% =====================================================================
% data source (Sisyphus recipe): utils.table_data.WriteTableDataJob (flat)
% per-row comments below map columns -> the registered output name the number comes from
\begin{table}[tb]
  \centering
  \caption{\textbf{Loquacious scaling:}
training on 250h to 25\,000h paired audio,
always injecting the full 25\,000h transcripts.
\textit{Used text : audio}: amount of unpaired text vs.~paired text;
\textit{Seen text : audio} over the training is about 1:1 for every injection run.}

  \label{tab:loq-scale}
  \begin{adjustbox}{max width=\linewidth}
  \begin{tabular}{|c|c|l|c|c|c|c|c|c|}
    \hline
    \multirow{2}{*}{\textbf{\makecell{Sub- \\ set}}} & \multirow{2}{*}{\textbf{\makecell{Audio \\ {\normalfont [h]}}}} & \multirow{2}{*}{\textbf{\makecell[l]{Text \\ injection}}} & \textbf{\makecell{Used}} & \multirow{2}{*}{\textbf{\makecell{Audio \\ pass.}}} & \textbf{\makecell{Up-}} & \multicolumn{2}{c|}{\textbf{WER {\normalfont [\%]}}} & \textbf{\makecell{Train}} \\
     &  &  & \textbf{\makecell{text : \\ audio}} &  & \textbf{\makecell{dates \\ {\normalfont [k]}}} & \textbf{dev} & \textbf{test} & \textbf{\makecell{time \\ {\normalfont [h]}}} \\ \hline\hline
    \multirow{4}{*}{\rotatebox{90}{small}} & \multirow{4}{*}{250} & \multirow{3}{*}{none} & \multirow{3}{*}{-} & \multirow{2}{*}{200} & $23$ & $15.60$ & $16.67$ & $5$ \\
    \cline{6-9}
     &  &  &  &  & $32$ & $15.45$ & $16.65$ & $5$ \\
    \cline{5-9}
     &  &  &  & 400 & $45$ & $16.34$ & $17.88$ & $10$ \\
    \cline{3-9}
     &  & \makecell[l]{mean \\ log-mel} & 100:1 & 200 & $64$ & $11.82$ & $13.02$ & $16$ \\
    \hline
    \multirow{4}{*}{\rotatebox{90}{medium1k}} & \multirow{4}{*}{1000} & \multirow{3}{*}{none} & \multirow{3}{*}{-} & \multirow{2}{*}{162} & $95$ & $8.04$ & $8.89$ & $19$ \\
    \cline{6-9}
     &  &  &  &  & $136$ & $7.82$ & $8.71$ & $22$ \\
    \cline{5-9}
     &  &  &  & 325 & $191$ & $8.07$ & $8.98$ & $39$ \\
    \cline{3-9}
     &  & \makecell[l]{mean \\ log-mel} & 25:1 & 162 & $209$ & $6.95$ & $7.81$ & $52$ \\
    \hline
    \multirow{4}{*}{\rotatebox{90}{medium}} & \multirow{4}{*}{2500} & \multirow{3}{*}{none} & \multirow{3}{*}{-} & \multirow{2}{*}{65} & $93$ & $6.88$ & $7.44$ & $19$ \\
    \cline{6-9}
     &  &  &  &  & $132$ & $6.60$ & $7.34$ & $21$ \\
    \cline{5-9}
     &  &  &  & 130 & $185$ & $6.41$ & $7.14$ & $38$ \\
    \cline{3-9}
     &  & \makecell[l]{mean \\ log-mel} & 10:1 & 65 & $133$ & $6.32$ & $7.24$ & $33$ \\
    \hline
    \multirow{3}{*}{\rotatebox{90}{large}} & \multirow{3}{*}{25000} & \multirow{2}{*}{none} & \multirow{2}{*}{-} & 2.8 & $95$ & $6.21$ & $6.90$ & $19$ \\
    \cline{5-9}
     &  &  &  & 5.6 & $191$ & $5.71$ & $6.29$ & $39$ \\
    \cline{3-9}
     &  & \makecell[l]{mean \\ log-mel} & 1:1 & 2.8 & $209$ & $6.00$ & $6.81$ & $52$ \\
    \hline
  \end{tabular}
  \end{adjustbox}
\end{table}

Despite the varying gain magnitudes across scales, the ablation trends on Loquacious confirm the LibriSpeech findings.
\Cref{tab:loq-text} and \Cref{tab:loq-representation} show that average WER is stable across text representations.
Just as on LibriSpeech, the exact acoustic features matter little:
mean log-mel features extracted from LibriSpeech perform almost identically to those extracted from Loquacious itself.
However, uniform sampling ($\alpha=0$) helps the YODAS subset, and more text helps the test set.
A second seed of the baseline and of the uniform-source injection reproduces the gain on the dev set (6.58 vs.\ 6.35), though test performance is flat (7.30 vs.\ 7.32).
However, a clear gain on both sets emerges when applying the Loquacious LM in search (CTC+AED+LM: 6.17 vs.\ 5.83 on dev, 6.97 vs.\ 6.73 on test).
% =====================================================================
% AUTO-GENERATED by scripts/render_tables.py -- DO NOT EDIT THIS FILE.
% Regenerate from (edit these, then re-run render_tables.py / sync_tables_preview.sh):
%   tables-data/loq-text.data.json    -- the numbers (from the Sisyphus graph)
%   tables-spec/loq-text.spec.json     -- columns, units, layout
%   tables-spec/loq-text.caption.tex   -- caption text
% =====================================================================
% data source (Sisyphus recipe): utils.table_data.WriteTableDataJob (flat)
% per-row comments below map columns -> the registered output name the number comes from
\begin{table}[tb]
  \centering
  \caption{\textbf{Loquacious data ratio and sampling:}
varying text amount
and source distribution
(sampled $\propto \mathrm{hours}^\alpha$, $\alpha = 1$: natural, $0$: uniform)
on the 2\,500h subset.
(VP: VoxPopuli, CV: Common Voice, LS: LibriSpeech, YO: YODAS.)}

  \label{tab:loq-text}
  \begin{adjustbox}{max width=\linewidth}
  \begin{tabular}{|c|c|c|c|c|c|c|c|c|}
    \hline
    \multirow{3}{*}{\textbf{$\alpha$}} & \multirow{3}{*}{\textbf{\makecell{Used \\ text : \\ audio}}} & \multirow{3}{*}{\textbf{\makecell{Seen \\ text : \\ audio}}} & \multicolumn{6}{c|}{\textbf{WER {\normalfont [\%]}}} \\
     &  &  & \multicolumn{5}{c|}{\textbf{dev}} & \multicolumn{1}{c|}{\textbf{test}} \\
     &  &  & \textbf{all} & \textbf{VP} & \textbf{CV} & \textbf{LS} & \textbf{YO} & \textbf{all} \\ \hline\hline
    - & 0:1 & 0:1 & $6.60$ & $6.45$ & $8.55$ & $4.79$ & $13.76$ & $7.34$ \\
    \hline
    \multirow{2}{*}{1} & \multirow{5}{*}{10:1} & 1:3.5 & $6.33$ & $6.07$ & $8.56$ & $4.55$ & $12.76$ & $7.26$ \\
    \cline{3-9}
     &  & 1:1 & $6.27$ & $5.92$ & $8.61$ & $4.43$ & $13.21$ & $7.05$ \\
    \cline{1-1}\cline{3-9}
    0.5 &  & \multirow{2}{*}{1:3.5} & $6.32$ & $6.14$ & $8.31$ & $4.62$ & $12.54$ & $7.15$ \\
    \cline{1-1}\cline{4-9}
    \multirow{2}{*}{0} &  &  & $6.32$ & $6.06$ & $8.43$ & $4.68$ & $11.98$ & $7.24$ \\
    \cline{3-9}
     &  & 1:1 & $6.33$ & $6.12$ & $8.21$ & $4.58$ & $13.81$ & $7.01$ \\
    \hline
  \end{tabular}
  \end{adjustbox}
\end{table}

% =====================================================================
% AUTO-GENERATED by scripts/render_tables.py -- DO NOT EDIT THIS FILE.
% Regenerate from (edit these, then re-run render_tables.py / sync_tables_preview.sh):
%   tables-data/loq-representation.data.json    -- the numbers (from the Sisyphus graph)
%   tables-spec/loq-representation.spec.json     -- columns, units, layout
%   tables-spec/loq-representation.caption.tex   -- caption text
% =====================================================================
% data source (Sisyphus recipe): utils.table_data.WriteTableDataJob (flat)
% per-row comments below map columns -> the registered output name the number comes from
\begin{table}[tb]
  \centering
  \caption{\textbf{Loquacious acoustic features:}
varying the acoustic features on the 2\,500h subset:
mean log-mel vectors from LibriSpeech or Loquacious,
textogram-style extra one-hot channels,
trainable log-mel embeddings.
}

  \label{tab:loq-representation}
  \begin{adjustbox}{max width=\linewidth}
  \begin{tabular}{|l|c|c|c|c|c|c|}
    \hline
    \multirow{3}{*}{\textbf{\makecell[l]{Acoustic \\ features}}} & \multicolumn{6}{c|}{\textbf{WER {\normalfont [\%]}}} \\
     & \multicolumn{5}{c|}{\textbf{dev}} & \multicolumn{1}{c|}{\textbf{test}} \\
     & \textbf{all} & \textbf{VP} & \textbf{CV} & \textbf{LS} & \textbf{YO} & \textbf{all} \\ \hline\hline
    no text & $6.60$ & $6.45$ & $8.55$ & $4.79$ & $13.76$ & $7.34$ \\
    \hline
    LS table & $6.32$ & $6.06$ & $8.43$ & $4.68$ & $11.98$ & $7.24$ \\
    \hline
    Loq. table & $6.44$ & $6.16$ & $8.52$ & $4.71$ & $13.18$ & $7.24$ \\
    \hline
    extra one-hot ch. & $6.41$ & $6.18$ & $8.50$ & $4.72$ & $12.45$ & $7.27$ \\
    \hline
    trainable log-mel & $6.40$ & $6.16$ & $8.52$ & $4.70$ & $12.47$ & $7.22$ \\
    \hline
  \end{tabular}
  \end{adjustbox}
\end{table}

\section{Conclusions}

A simple lookup table with durations from a phoneme alignment
can replace neural TTS for text injection.
Training is 1.8 times faster and
requires no synthesizer.
On LibriSpeech, it yields a 12\,\% relative WER reduction over the baseline,
reaching 1.36\,/\,2.90\,/\,1.52\,/\,3.24\,\% with a large model.
The duration statistics are the most critical component,
while the specific acoustic features matter less.
The best results rely on a phoneme lexicon.
While characters retain 60\,\% of the gain without one, generating robust inputs directly from subword alignments requires further investigation.
Finally, scaling experiments on Loquacious confirm these trends:
the overall relative gains scale predictably with the text-to-audio ratio and paired data amount,
the method is robust, and acoustic features generalize across datasets without retraining.
Comparing relative gains across the literature (\Cref{fig:ratio-gain}) suggests a clear trend:
at equal scale, injecting text at the input to train the full encoder generally outperforms injecting into a shared upper encoder or the decoder alone.

% 9 pt is the floor for every font in the paper, so no smaller type here.
% https://cmsworkshops.com/ICASSP2027/papers.php
% "An optional 5th page may be included, but that page may contain only references, funding acknowledgments, and a Compliance with Ethical Standards statement."

\section{Acknowledgements}

This work was partially supported by NeuroSys,
which as part of the initiative “Clusters4Future” is funded by the Federal Ministry of
Education and Research BMBF (funding IDs 03ZU2106DA and 03ZU2106DD),
and by the project RESCALE within the program
\textit{AI Lighthouse Projects for the Environment, Climate, Nature and Resources}
funded by
the Federal Ministry for the Environment, Nature Conservation,
Nuclear Safety and Consumer Protection (BMUV),
funding ID: 67KI32006A.
The authors gratefully acknowledge the computing time provided to them
at the NHR Center NHR4CES at RWTH Aachen University
(project number p0023565 and p0023999).
This is funded by the Federal Ministry of Education and Research,
and the state governments participating on the basis
of the resolutions of the GWK for
national high performance computing at universities
(\href{www.nhr-verein.de/unsere-partner}{www.nhr-verein.de/unsere-partner}).

% no \vfill\pagebreak before the references: in two columns it only breaks the column,
% and the 4th page may hold references anyway (the kit)

\bibliographystyle{IEEEbib}
\bibliography{strings,refs}

\end{document}